%% file: root.tex
\documentclass[letterpaper, 10 pt, conference]{ieeeconf}  

\IEEEoverridecommandlockouts                              
\usepackage{cite}
\usepackage{amsmath,amssymb,amsfonts}
\usepackage{graphicx}
\usepackage{textcomp}
\usepackage{xcolor}
\def\BibTeX{{\rm B\kern-.05em{\sc i\kern-.025em b}\kern-.08em
    T\kern-.1667em\lower.7ex\hbox{E}\kern-.125emX}}
\usepackage{hyperref}
\let\labelindent\relax
\usepackage{enumitem}
\usepackage{multirow}
\usepackage{xspace}
\usepackage{vertbars}
\usepackage{tikz}
\usepackage{lscape, float}
\usepackage{rotating}
\usepackage[normalem]{ulem}
\usepackage{placeins}
\usepackage{bm}
\usepackage{balance} 
\usepackage{algorithm}
\usepackage{algpseudocode}
\usepackage{tcolorbox}
\usepackage{comment}
\usepackage{xspace}

\newcommand{\frameworkName}[0]{BT-ACTION\xspace}

\title{\LARGE \bf
\frameworkName: A Test-Driven Approach  for Modular Understanding of User Instruction Leveraging Behaviour Trees and LLMs}

\author{Alexander Leszczynski$^{1}$, Sarah Gillet$^{1}$, Iolanda Leite$^{1}$  and Fethiye Irmak Dogan$^{1}$%
\thanks{$^{1}$ A. Leszczynski, S. Gillet, I. Leite, and F. I. Dogan are with the Division of Robotics, Perception and Learning, KTH Royal Institute of Technology, Stockholm, Sweden. {\tt\small \{aleles, sgillet, iolanda, fidogan\}@kth.se.}}%
}

\begin{document}

\maketitle
\thispagestyle{empty}
\pagestyle{empty}

\begin{abstract}
\input{sections/00_abstract}
\end{abstract}

\input{sections/01_introduction}
\input{sections/02_related_work}

\input{sections/method_final}
\input{sections/evaluation2}

\input{sections/05_discussion}
\bibliographystyle{IEEEtran}
\bibliography{references_tidy}

\end{document}

%% file: sections/00_abstract.tex
Natural language instructions are often abstract and complex, requiring robots to execute multiple subtasks even for seemingly simple queries. For example, when a user asks a robot to prepare avocado toast, the task involves several sequential steps. Moreover, such instructions can be ambiguous or infeasible for the robot or may exceed the robot's existing knowledge. While Large Language Models (LLMs) offer strong language reasoning capabilities to handle these challenges, effectively integrating them into robotic systems remains a key challenge. To address this, we propose BT-ACTION, a test-driven approach that combines the modular structure of Behavior Trees (BT) with LLMs to generate coherent sequences of robot actions for following complex user instructions, specifically in the context of preparing recipes in a kitchen-assistance setting. We evaluated BT-ACTION in a comprehensive user study with 45 participants, comparing its performance to direct LLM prompting. Results demonstrate that the modular design of BT-ACTION helped the robot make fewer mistakes and increased user trust, and participants showed a significant preference for the robot leveraging BT-ACTION. The code is publicly available at \href{https://github.com/1Eggbert7/BT_LLM}{https://github.com/1Eggbert7/BT\_LLM}.

%% file: sections/01_introduction.tex
\section{INTRODUCTION}


Following user instructions presents several challenges for robots operating in real-world environments, as natural language commands are often abstract and typically involve multiple sub-tasks that the robot must identify and execute~\cite{Arumugam-RSS-17}. Even seemingly simple commands, such as \textit{``Prepare me an avocado toast''}, require a series of actions, including toasting the bread, peeling the avocado, and serving the final dish on a plate. Moreover, such instructions may be ambiguous (e.g., whether additional ingredients should be included), infeasible (e.g., no bread available), or outside the robot's current knowledge base (e.g., unfamiliar variations like ham toast). In these cases, modular approaches that decompose tasks into manageable sub-tasks can help robots complete instructions more effectively \cite{meng_dcr_2024}.

Prior work has explored various methods to improve how robots interpret and act upon human instructions in diverse environments~\cite{kilicaslan2013nlp, cantrell2010robust, paxton_costar_2016, dogan_asking_2022, dogan2024semantically}. Recent advancements in Large Language Models (LLMs) have opened new opportunities to enhance these interactions~\cite{stilinki_bridging_2024, kim_understanding_2024, wang_wall-e_2023, wang_lami_2024}, particularly in identifying uncertainty and ambiguity~\cite{wang_wall-e_2023, park_clara_2024}. To support effective comprehension of user instructions, step-by-step approaches that decompose complex tasks into subtasks have been employed~\cite{10011755, Arumugam-RSS-17}, with Behavior Trees (BTs) emerging as a promising structure for managing ambiguous queries~\cite{iovino_survey_2022}.
While these studies have laid important groundwork, the challenge remains: How can robotic systems leverage the reasoning power of LLMs while maintaining a modular, adaptable structure that evolves with novel queries using prior robot knowledge? To address this, \textit{\textbf{we propose a test-driven approach that resolves tasks step by step, utilizing the modular structure of BTs and using LLMs in each step to handle ambiguities, communicate the robot’s capabilities, and generalize to novel user requests}}.

In this paper, we introduce \frameworkName, a hierarchical test-driven approach for generating high-level actions (e.g., first \texttt{cook\_egg}, then \texttt{place\_on\_sandwich}) that enable a robot to follow complex user queries in a kitchen-assistance setup. \frameworkName first constructs a BT that classifies user instructions into categories, i.e., ambiguous, clear, infeasible, or modification requests, the latter referring to queries that go beyond the robot’s prior knowledge but are still executable. Each of these instruction types is then processed using LLMs, enabling nuanced interpretation and response. 
The modular design of \frameworkName allows the robot to communicate its own capabilities, handle errors, and highlight the sources of ambiguities in user instructions. To evaluate the effectiveness of our approach, we conducted an in-person user study with 45 participants who interacted with a Furhat robot by requesting to fulfill several tasks and prepare various recipes. The results show that the robot made significantly fewer errors (exhibiting reduced hallucinations and incorrect action steps) and enhanced user trust, with participants showing a strong preference when the robot leveraged the modular architecture of \frameworkName compared to direct LLM prompting.

%% file: sections/02_related_work.tex
\section{RELATED WORK}


Comprehending user instructions has been a critical focus in human-robot interaction (HRI) research, particularly in developing robust systems for effective verbal communication~\cite{kilicaslan2013nlp, cantrell2010robust, paxton_costar_2016}. To tackle this challenge, many studies have explored informative clarification strategies for handling ambiguous or underspecified user queries~\cite{dogan_asking_2022, shridhar2020ingress, yang2022interactive, mees2020composing}. For example, Dogan et al.~\cite{dogan_asking_2022, dogan_leveraging_2023} have identified sources of ambiguity in user instructions through explainability techniques, while Yang et al.~\cite{yang2022interactive} have employed attribute-based disambiguation methods to improve object grasping in HRI.


Recent work on understanding user instructions has increasingly leveraged the power of Large Language Models (LLMs) to process natural language input and generate contextually appropriate responses~\cite{stilinki_bridging_2024, kim_understanding_2024, wang_wall-e_2023, wang_lami_2024}. For example, Wang et al.~\cite{wang_wall-e_2023} have employed LLMs for interactive disambiguation, followed by visual grounding to identify objects referenced in user queries. A notable recent approach, CLARA~\cite{park_clara_2024}, has used an LLM to classify user instructions as clear, ambiguous, or infeasible. In contrast, our work proposes a modular approach that not only performs such classifications but also generates a structured sequence of high-level actions, enabling robots to support end-to-end task execution grounded in user commands.


Prior research has shown that breaking down complex tasks into manageable units improves both performance and system reliability~\cite{holk_predilect_2024, meng_dcr_2024}. In parallel to our work, other studies have focused on decomposing complex user queries into simpler, actionable components~\cite{10011755, Arumugam-RSS-17}. More closely related, Iovino et al.~\cite{iovino_interactive_2022} employed behavior trees (BTs) to manage ambiguous instructions in tabletop scenarios involving robotic object search. While these efforts provide a strong foundation for task decomposition, they do not leverage a robot's existing knowledge to adapt to novel user queries that are feasible but require modification of an existing knowledge set. Moreover, they lack mechanisms to inform users about requests outside of robot capabilities while generating step-by-step actionable units. \frameworkName addresses these limitations with a test-driven approach by first classifying user queries and then employing LLMs at different BT nodes to provide explanations and support interactive clarification during task execution.

%% file: sections/method_final.tex
\section{Methodology: \frameworkName}

\label{ch:methodology}

\subsection{Overview of the Framework}

\begin{figure}[t!]
    \centering
    \includegraphics[width=0.8\linewidth]{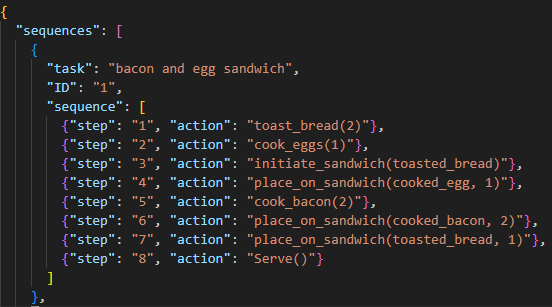}
    \vspace{-0.5em}
    \caption{An example set of actions for task \( T_i\) (preparing bacon and egg sandwich), each composed of step-by-step high-level robot actions, \( T_i = \{a_0,...,a_8\}\),  \(a_j \in A \)}
    \label{fig:Seq-example}
    \vspace{-1.5em}
\end{figure}

Given a user instruction \( u \), the goal of \frameworkName is to generate a set of robot actions that would fulfill the task specified in \( u \).
Accordingly, each task \( T_i \) is composed of a set of \(a_j \in A\) that should be executed step by step, \( T_i~=~\{a_0 \dots a_j\} \) (see Figure~\ref{fig:Seq-example}). Moreover, \(\mathcal{T}~=~\{T_0 \dots T_n\} \) shows the set of tasks that the robot is familiar with.

\begin{figure*}[t]
    \centering
    \includegraphics[width=0.9\linewidth]{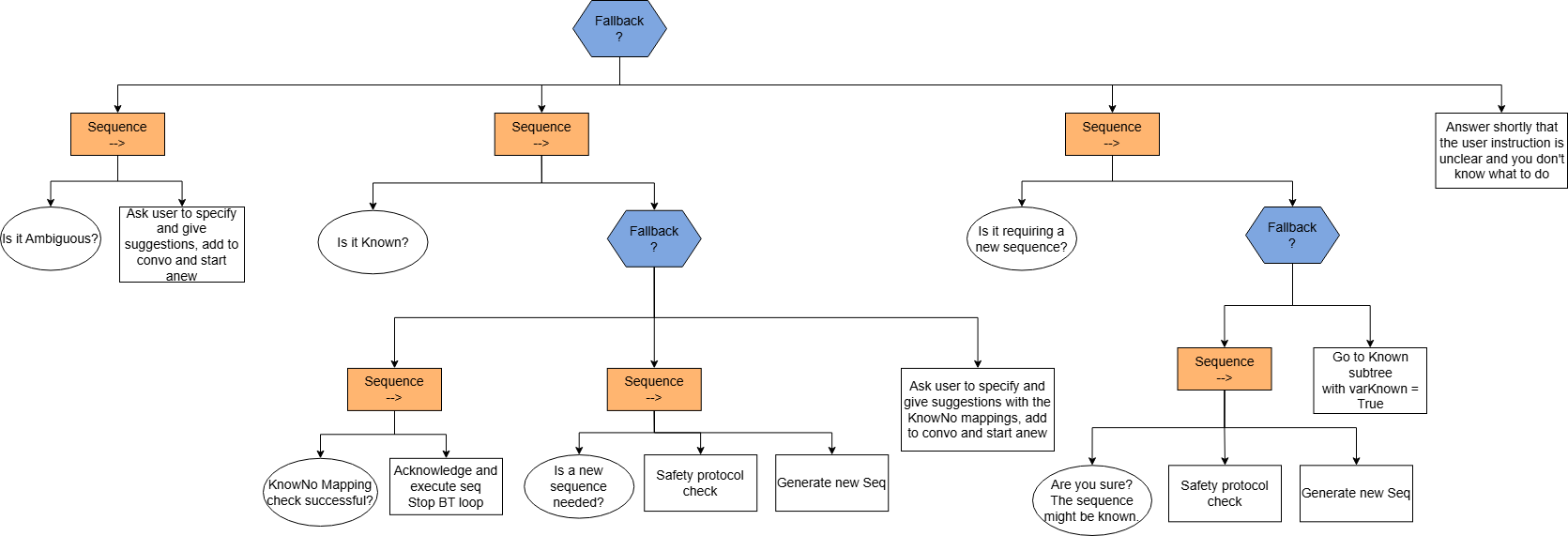}
        \vspace{-0.5em}
    \caption{Simplified Behavior Tree of the \frameworkName System.}
    \label{fig:simplified_tree}
    \vspace{-1.5em}
\end{figure*}

To comprehend and act upon the user instructions, \frameworkName follows a test-driven approach and suggests four main components to construct the behavior tree: Classification, Information, Sequence generation, and Error Handling. Classification enables more fine-tuned handling of user instructions through LLM-based condition nodes; information ensures explanations and transparency in robot behavior; sequence generation creates a new set of actions for novel tasks; and error handling solidifies the robustness of the framework by checking for potential mistakes and LLM hallucinations. A simplified version of the \frameworkName can be seen in Figure~\ref{fig:simplified_tree}.

\subsubsection{Classification}
\label{class}
\frameworkName leverages the modular structure of BTs to systematically process and classify user instructions $u$ into the following categories:

\[\footnotesize
C(u) =
\begin{cases} 
\texttt{Clear}, & \text{if it directly maps to a known sequence,} \\
\texttt{Ambiguous}, & \text{if clarification is required,} \\
\texttt{Modification}, & \text{if modification of a known task is needed,} \\
\texttt{Infeasible}, & \text{if beyond the robots capabilities.}
\end{cases}
\]

This classification is implemented through a sequence of condition nodes in the behavior tree, each using a tailored LLM pre-prompt, for example: \textit{``Based on the user instruction and history, is the request ambiguous?''} or \textit{``Does the instruction match a known task''}. This classification starts with ambiguity detection and is followed by checks for known task matches and feasibility. Concretely, classification proceeds through a stepwise series of binary checks:
\begin{equation}
   C(u) = \texttt{LLM}(f(u) \oplus g(H)),
\end{equation}
\noindent where \( f(u) \) embeds the user instruction into prompt-ready format, \( g(H) \) encodes the current conversation history, and $ \oplus$ shows the string concatenation operator to form the prompt. Each check corresponds to a condition node in the behavior tree, and the first one returning true determines the classification and activates the corresponding information subtree.

\subsubsection{Information} \label{sec:clarifciation}
\label{sec:information}
After classification, the robot’s goal is to communicate its intended actions back to the user based on the classification category:

(a) If \( C(u) = \texttt{Clear} \), the framework first identifies the matching task \(T_i \in \mathcal{T}, i \in \{1,...,n\} \) from the set of known tasks. This match is determined through \texttt{KNOWNO} error-handling (see Section~\ref{error}). Once validated, the robot informs users about the execution of task $T_i$, such as \textit{``I'll get started on the task for a bacon and egg sandwich''}.

(b) If \( C(u) = \texttt{Ambiguous} \), the LLM is prompted to generate a follow-up question: 
\(
Q~\leftarrow~\texttt{LLM}(g(H)~\oplus~\mathcal{T}~\oplus~P_{\text{follow-up}}),
\)
where \( g(H) \) encodes the full conversation history $H$, \( \mathcal{T} = \{T_0, T_1, ..., T_n\} \) is the set of known tasks, and \( P_{\text{follow-up}} \) is a structured pre-prompt instructing the LLM to suggest known tasks based on the conversation history and to inquire if the user has something else in mind. Hence, $Q$ includes possible matching options, e.g., \textit{``There are multiple options that might fit your request. Would you like some pancakes with maple syrup and berries or perhaps a peanut butter and jelly sandwich?''}. The user's reply \( r \) is appended to the conversation history \( H \leftarrow H \cup \{(Q, r)\} \), and the behavior tree is re-triggered from the root with the updated content.

(c) If \( C(u) = \texttt{Modification} \), the system triggers the sequence generation process (see Section~\ref{sec:seqgen}) to obtain a new task \( T_{n+1} \). Then, \( T_{n+1} \) is validated with the \texttt{CheckNewSeq} error-handling mechanism to avoid hallucinations (see Section~\ref{error}). If verified, LLM is prompted to summarize \( T_{n+1} \) for further user confirmation, $\texttt{LLM}(T_{n+1})$, such as \textit{``To fulfill your request, I will fry two pieces of bacon and an egg, then put it on a toast and serve. Does this sound good to you?''}. Finally, if the user confirms the provided \( T_{n+1} \), the sequence is accepted as the result of the interaction with the user; if not, the robot asks users to restate their instructions in a clearer way. Then, the user's response is appended to the history $H$, and the tree is re-triggered from the root.

(d) If \( C(u) = \texttt{Infeasible} \), the robot informs the user that their request cannot be fulfilled due to resource constraints or limitations in its capabilities. The robot communicates an appropriate explanation using the LLM:
\(
\texttt{LLM}(u, A, I, P_{\text{explanation}}),
\)
where \( I \) is the set of existing ingredients, and \( P_{\text{explanation}} \) is a structured prompt guiding the LLM to explain why the task is infeasible. Example responses might include: \textit{``I’m sorry, but I cannot repaint the kitchen walls as I lack the necessary tools''}
or
\textit{``I cannot make this request as it includes ‘goat cheese’, which is not part of existing ingredients''.}
The robot's response \( r \) is appended to the conversation history:
\(
H \leftarrow H \cup \{(u, r)\}
\),
and the behavior tree is re-triggered from the root with the updated content to handle the cases where users want to redefine their requests.

\subsubsection{Sequence generation}
\label{sec:seqgen}
When a user instruction is classified as a \texttt{Modification}, the system prompts the LLM to generate a new task action set $T_{n+1}$: 
\begin{equation}
T_{t+1} \leftarrow \texttt{LLM} (f(u) \oplus g(H) \oplus w(\mathcal{T})),
\end{equation}
where \( f(u) \) embeds the instruction into prompt-ready format, \( g(H) \) encodes the full conversation history, and \( w(\mathcal{T}) \) lists all known tasks \( \{T_0,\dots, T_n\} \).
The output \( T_{n+1} \) is the set of robot actions to fulfill the query $u$. An example output of such a generated sequence can be seen in Figure~\ref{fig:Seq-example}.

\subsubsection{Error Handling}
\label{error}

\frameworkName also incorporates the following error-handling to further support robustness and avoid LLM hallucinations:

(a) \texttt{KNOWNO:} 
 This safety check is integrated for cases labeled as \texttt{Clear} to ensure the user instruction only maps one known task. It uses the KNOWNO framework~\cite{ren_robots_2023}, which measures the uncertainty of LLM-based planners. 
KNOWNO takes the user request \( u \) as input and returns a set of tasks \( T = \{T_0,\dots, T_n\} \) that potentially fulfill the request. If multiple mappings are found, the robot provides these options to the user for further clarification as in the \texttt{Ambigious} cases. Differently, this node does not use LLMs to provide these options but returns the ones obtained from the KNOWNO framework.

(b) \texttt{CheckNewSeq:}
In the case of a \texttt{Modification} request, the LLM-generated sequence $T_{n+1}$ is parsed to ensure all actions are part of the predefined set of valid robot actions \( A = \{a_0,\dots, a_n\} \), and all mentioned ingredients exist within the existing ingredient list \( I = \{i_0,\dots, i_m\} \), aiming to avoid LLM hallucinations.
An additional safety check for \texttt{Modification} cases is to assess whether the request is feasible in terms of available ingredients and reasonable quantities. To archive this, the LLM is prompted with the encoded ingredient list \( I \) and defining acceptable quantity limits (e.g., under 10 units). 
If any constraints on robot actions, ingredients or quantities are violated, the classification is changed as \texttt{Infeasible}, and the response is generated following the procedure in Section~\ref{sec:information}).

\subsection{Implementation Details}
\label{sec:softwareDevelopment}
An overview of how a user instruction is processed through \frameworkName is summarised in Algorithm~\ref{alg:btplan}. The \frameworkName was implemented using \texttt{PyTrees\footnote{https://py-trees.readthedocs.io/en/devel/}} and used the OpenAI ChatGPT 4o-mini for LLM prompting.

The modular design allowed each node—e.g., ambiguity classification or feasibility checking—to be developed and tested independently before integration. This structure aimed to ensure that system behavior remained explainable and predictable, reinforcing the goal of structured and context-aware user instruction understanding.

\subsection{Baseline System} \label{sec:baseline}
The baseline system served as a comparison for evaluating the performance of \frameworkName. For each user instruction \( u \), the base prompt was constructed as:
\(
y \leftarrow \texttt{LLM}(u, S, P_{\text{base}})
\)
where \( u \) is the user instruction, \( S \) is the system context, including all known task sequences, available ingredients, and high-level actions, \( P_{\text{base}} \) is a long static prompt guiding the LLM to interpret, clarify, and respond to user requests as needed. 

Specifically, \( P_{\text{base}} \) hanleded the following cases: (i)~$\texttt{Clear}:$ \textit{``For requests
that exactly match a sequence above,
acknowledge and indicate you’ll commence the
task~$\dots$''}, (ii)~$\texttt{Ambiguous}:$ \textit{``If a request is not clear or detailed enough; ask the user to be more specific~$\dots$''}, (iii)~$\texttt{Modification}:$ \textit{``For requests
that are slightly different from your
pre-programmed tasks, first check if you can
generate a new sequence by strictly using the defined functions and ingredients~$\dots$''}, (iv)~$\texttt{Infeasible}:$ \textit{``If a
request falls entirely outside of your
capabilities, decline politely and explain your limitations~$\dots$''}.

The base system relied on a single static pre-prompt loaded into the ChatGPT-4o-mini API. Hence, it was designed to generate responses without the support of a behavior tree and did not include BT error-handling mechanisms.

\begin{algorithm}[t!]
\caption{Behavior Tree Execution for \frameworkName.}
\label{alg:btplan}
\begin{algorithmic}[1]
\footnotesize
\Statex{$u$: User instruction, $H$:  Conversation history, $C(u)$: Instruction classifier ($\texttt{Ambiguous}$, $\texttt{Clear}$, $\texttt{Modification}$, or $\texttt{Infeasible}$)}
\vspace{0.5em}\If{$C(u) = \texttt{Clear}$}
    \State \texttt{KNOWNO} check to ensure one known mapping exists
    \State Acknowledge users about the sequence execution
\vspace{0.5em}\ElsIf{$C(u) = \texttt{Ambiguous}$}
    \State $Q~\leftarrow~\texttt{LLM}(g(H)~\oplus~\mathcal{T}~\oplus~P_{\text{follow-up}})$
    \State Obtain users users's reply $r$ to question $Q$
    \State \( H \leftarrow H \cup \{(Q, r)\} \)
    \State Re-trigger the behaviour tree from root
\vspace{0.5em}\ElsIf{$C(u) = \texttt{Modification}$}
    \State $p \leftarrow f(u) \oplus g(H) \oplus w(T)$
    \State $T_{n+1} \leftarrow \texttt{LLM}(p)$
    \State Confirm $T_{n+1}$ with user
    \State Execute $T_{n+1}$, acknowledge users
\vspace{0.5em}\ElsIf{$C(u) = \texttt{Infeasible}$}
    \State $\texttt{LLM}(u, A, I, P_{\text{explanation}})$
    \State Explain why the task is infeasible
    \State Re-trigger the behaviour tree from root
\EndIf
\end{algorithmic}
\end{algorithm}

%% file: sections/evaluation2.tex
\section{EVALUATION}
\subsection{Evaluation on case dataset} \label{sec:preevaluation}
To enable test-driven development, we start with simple LLM prompts in both the baseline system and \frameworkName, refining them using a case dataset. This case dataset contained 18 example user requests from different categories outlined in \ref{class}. An excerpt of this dataset is given in Table \ref{tab:test_set_feasible}. We continually refined the baseline prompt until further refinements did not improve the LLM output.  Then, we repeated the same process for \frameworkName. Therefore, we developed both systems to achieve high correctness in the case dataset for a fair comparison. Eventually, both the baseline prompt and \frameworkName yielded similar performances, where the baseline system got 16 of 18 cases correct and \frameworkName was correct in 17 out of 18 cases.


\begin{table}[t]
    \centering
    \caption{One example for each type out of the 18 total examples in the "Desired Behaviors"-data set.}
    \begin{tabular}{|p{0.075\textwidth}|p{0.15\textwidth}|p{0.175\textwidth}|}
        \hline
        \footnotesize
         \textbf{Case} & \textbf{User Instruction} & \textbf{Expected Behavior} \\
        \hline
        Clear & Can I get the bacon and egg sandwich? & Direct execution of the known bacon and egg sandwich sequence without further clarification. \\
        \hline
        Ambiguous & I am hungry, can I have something to eat? & Prompt for more specific instructions or suggest generating a new sequence based on user preference. \\
        \hline
        Modification & Make me pancakes but without the berries and double the amount of maple syrup. & Generate a new sequence by combining parts of known sequences to fulfill the request. \\
        \hline
        Infeasible & Please repaint the kitchen walls. & Inform the user that the task is infeasible due to limitations in capabilities or resources. \\
        \hline
    \end{tabular}
    \label{tab:test_set_feasible}
    \vspace{-2em}
\end{table}


\subsection{Evaluation in user study}\label{sec:user-study}
We designed a within-participant user study with two conditions to evaluate the effectiveness of \frameworkName in comparison to the baseline system in the kitchen-assistant scenario and formulated the following hypotheses:

\begin{itemize}
    \item[\textbf{H1:}] As decomposing complex tasks into manageable sub-units has been suggested to enhance system capabilities~\cite{meng_dcr_2024}, we hypothesized that \frameworkName would reduce task errors compared to the baseline system.
    \item[\textbf{H2:}] As predictable robotic behavior has been perceived as more competent~\cite{10.1145/3461534}, we hypothesized that the robot using \frameworkName will be perceived as more competent, warmer, and less discomforting than the robot using baseline system. 
    \item[\textbf{H3:}] As reduced task errors have been shown to increase trust towards robots~\cite {doi:10.1177/0018720811417254}, we hypothesized that participants would rate their trust toward the robot using \frameworkName higher than the baseline and find the robot using \frameworkName more explainable.
    \item[\textbf{H4:}] As robots with clear and informative feedback systems preferred by people~\cite{lim_why_2009}, we hypothesized that participants would prefer the robot using \frameworkName over the baseline system.
\end{itemize}

\subsubsection{Conditions}
We used a test-driven development approach to design two different conditions in which the robot can act as a kitchen assistant:

\textbf{\frameworkName Condition:} The \frameworkName condition implemented the system as described in Section \ref{ch:methodology}.

\textbf{Control Condition:}
The robot in the baseline condition used the baseline prompt as described in Section \ref{sec:baseline}.

\subsubsection{Measures}


We used a mix of objective and subjective measures to understand the performance of the baseline and the \frameworkName. These measures aim to capture both the technical performance of the systems and the participants' experiences.

Objective measures included time measurements, turn count in conversation and task errors.
\begin{description}[align=left,leftmargin=0em,labelsep=0.2em,font=\textbf,itemsep=0em,parsep=0.3em]
\item{\textbf{Time Measurement}:}
The time taken for each interaction between the participant and the system was recorded at multiple points: the start of the interaction, at each conversational turn, and at the end of the interaction.

\item{\textbf{Turn Count}:}
The number of turns in the conversation was tracked by incrementing a turn counter after each exchange between the robot and the participant.

\item{\textbf{Number of Task Errors}:}
Task errors were defined as deviations from the expected behavior or response in the context of the predefined tasks. We categorized errors observed during the user study into six types: Lie/Hallucination, Faulty JSON, False Execution, Unnecessary JSON Generation, Presumptive Execution, and Misclassification. We counted their occurrences for both systems.

\begin{itemize}
    \item \textit{Lie/Hallucination} errors occur when the system provides fabricated or incorrect information unrelated to the user’s request or the system’s capabilities. For example, incorrectly suggesting ingredients which are not part of the available ingredients. 


    

\item \textit{Faulty output} (JSON format) errors occur when the structured output format (e.g., see Figure~\ref{fig:Seq-example} for JSON output structure) used for task execution contains mistakes, e.g. mismatched quantities of ingredients for cooking and serving.

\item \textit{False Execution} describes errors in which the system wants to execute a wrong action. 

\item \textit{Unnecessary JSON Generation} describes errors in which the system unnecessarily generates JSON outputs, even when no new sequence is required.

\item \textit{Presumptive execution} occurs when the system assumes the user’s intent in ambiguous requests, resulting in premature or incorrect actions.

\item \textit{Misclassification} is unique to the BT framework, which happens when the system misclassifies a known sequence as unknown, unnecessarily triggering sequence generation or providing an ambiguous response that does not lead to action execution. 
\end{itemize}

\end{description}

\begin{figure}[t]
    \centering
    \includegraphics[width=0.3\textwidth]{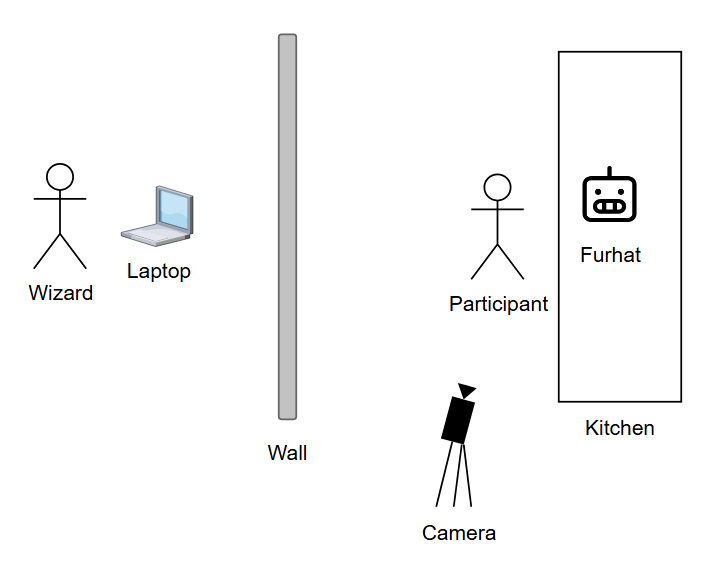}
    \vspace{-0.5em}
    \caption{Sketch of the Experiment setup.}
    \label{fig:experiment-setup}
    \vspace{-2em}
\end{figure}

\label{sec:measures}


We used subjective measures to evaluate participants’ perceptions of the robot, their trust toward the robot and their perception of the clarity of its actions.  
\begin{description}
[align=left,leftmargin=0em,labelsep=0.2em,font=\textbf,itemsep=0em,parsep=0.3em]
\item{\textbf{Perception of the robot:}}
To measure participants' perception of the robot, we use the RoSAS questionnaire~\cite{carpinella_robotic_2017}. Particularly, RoSAS measures the perceived warmth, competence and discomfort of the robot.

\begin{figure}[t]
    \centering
    \includegraphics[width=0.35\textwidth]{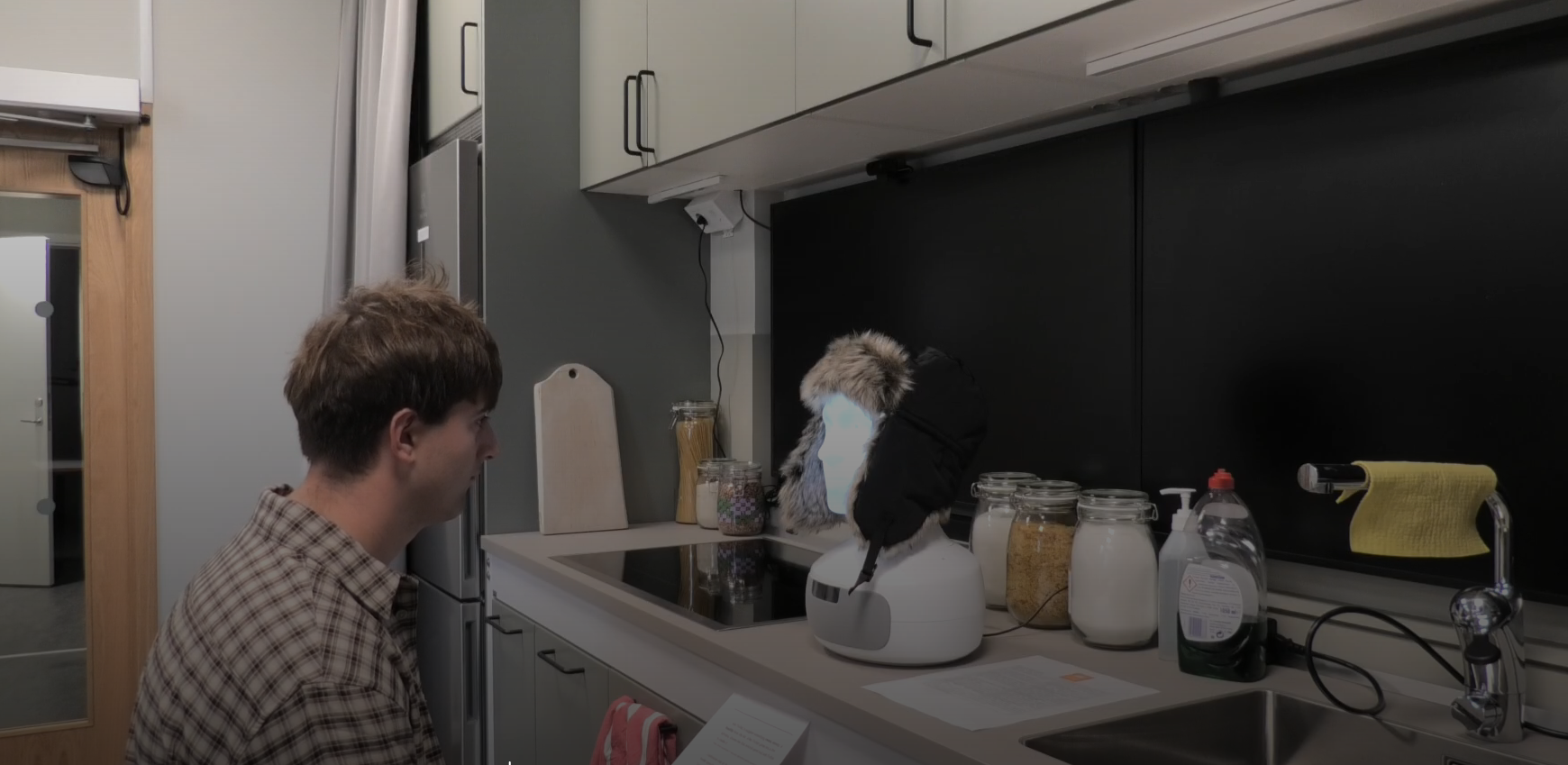}
    \caption{Picture from the experiment setup.}
    \label{fig:experiment-setup-pic}
    \vspace{-2em}
\end{figure}

\item{\textbf{Trust:}} 
We use a Multi-Dimensional Measure of Trust (MDMT)~\cite{ullman_measuring_2019} to evaluate participant's \textit{Capacity Trust} and \textit{Moral Trust} toward the robot.

\item{\textbf{Explainability:}}
 We use the explainability questionnaire designed by Silva et al.~\cite{silva_explainable_2023} to understand participants' perceptions of the robot’s decision-making processes and the clarity of its actions.

\item{\textbf{System preference:}}
In addition to these questionnaires, we asked participants which interaction they preferred. 
\end{description}

\subsubsection{Procedure}
After giving informed consent, participants were asked to fill out a demographic survey and were introduced to the context of the experiment: the Furhat robot acting as their home assistant manager (see Figure~\ref{fig:experiment-setup} and Figure~\ref{fig:experiment-setup-pic} for the experiment setup). Each participant was provided with a task sheet outlining five tasks to be completed during the session and a menu with different food options. We designed the five tasks in order to evaluate the systems' ability to handle user instructions across various scenarios, including ambiguous and complex inputs. We asked them to: ``Order an item as it is from the menu'', ``Order an item with a variation from the menu'', ``Ask for a rough category like sweet or saviour'', ``Order something that is not on the menu'', ``Order anything they want''.

The experiment began with participants interacting with the first system (counterbalanced). We asked participants to go through each of the five tasks. Upon completing the five tasks, participants were asked to fill out the questionnaire on their subjective perception of their experience. While participants filled out the questionnaire in a separate area where they could not see the robot, the experimenter prepared the second system. To give the impression that participants were interacting with a different robot, the experimenter adjusted its hat (either added it or removed it, counterbalanced). Participants then proceeded to interact with the second system using the same task sheet. Following this interaction, they completed a second survey to rate their experience with the second system.

At the end of the session, the participants were asked about their system preference. The experimenter concluded the session with a debriefing. Participants received a 100 SEK ($\approx$10 dollar) gift card as compensation for their time.

\begin{figure}[!t]
  \centering
  \includegraphics[width=0.35\textwidth]{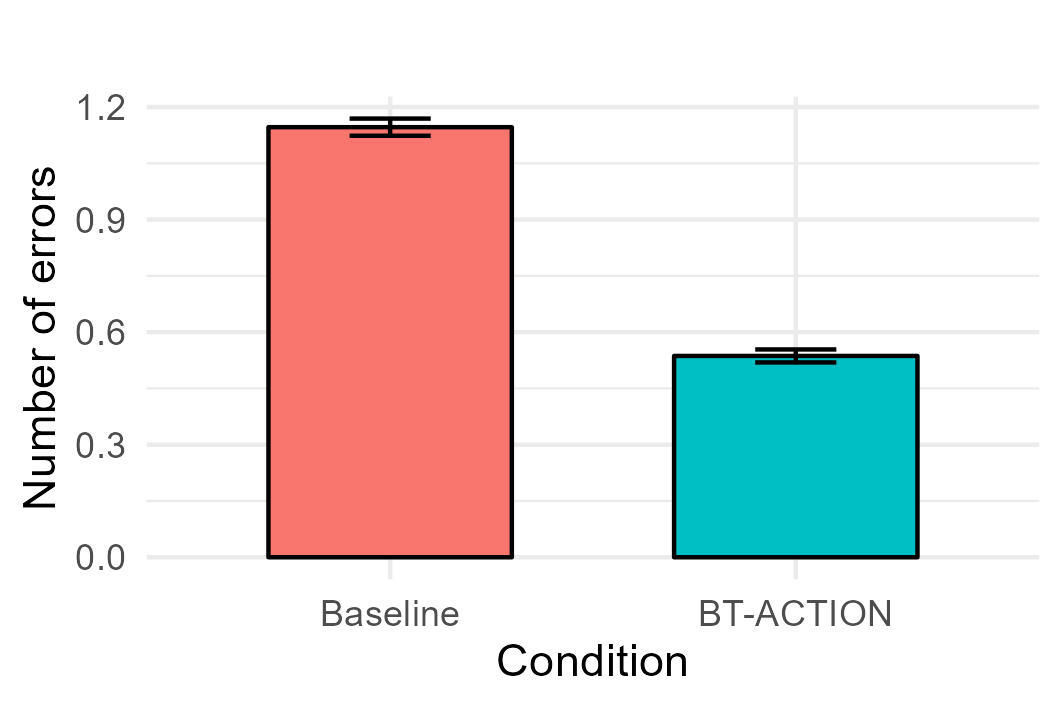}
    \vspace{-0.5em}
  \caption{Mean and standard error of the number of mistakes for the two conditions.}
  \label{fig:ViolinPlot}
\end{figure}

\subsubsection{Participants}
In total, we recruited 45 participants older than 18 and with English proficiency from the university campus and the surrounding town through online platforms, personal invitations, user study mailing lists, and flyers. 
The target participant size was determined using a power analysis for paired designs~\cite{bartlett_have_2022, bethel_review_2010}, considering an expected moderate effect size (Cohen's $d = 0.5$)~\cite{cohen_statistical_1988}, a significance level of $\alpha = 0.05$, and a power of $0.80$. This analysis suggested a minimum of 34 participants to detect meaningful differences, with additional participants recruited to account for potential dropouts or unusable data. The final participant size exceeded this minimum threshold.

\subsubsection{Results}
Of the 45 participants, four had to be excluded due to technical issues such as disrupted API calls and internet disruptions. The results reported for the remaining 41 participants, 53.7\% of the identified as male and 46.3\% as female with ages ranging from 20 to 79 (M = 28.6 years, STD = 10.4 years).  


\begin{table}[t]
\centering
\caption{Mean and Standard deviation for the subscales of the RoSaS questionnaire.}
\label{tab:descriptive_stats_rosas}
\begin{tabular}{lccc}
\textbf{System} & \textbf{Warmth} & \textbf{Competence} & \textbf{Discomfort} \\
Baseline & 3.50 (1.28) & 4.76 (1.11) & 2.21 (0.99) \\
\frameworkName & 3.70 (1.12) & 5.02 (1.01) & 1.91 (0.86) \\
\end{tabular}
\vspace{-2em}
\end{table}
\paragraph{Task Errors}
\label{sec:hypothesisH1}

We compared the number of task errors made by \frameworkName and the baseline system illustrated in Figure~\ref{fig:ViolinPlot}. 
A Shapiro-Wilk test indicated that the mistake data was non-normally distributed. Therefore, we used a Wilcoxon Signed Rank Test, which revealed a statistically significant difference between \frameworkName and the baseline (\( V = 331, p = 0.0025 \)) with \frameworkName (M=0.54, SD=0.71) making signficantly fewer mistakes than the baseline ( M=1.15, SD=0.94). Cliff’s Delta (\( \delta = 0.388 \), 95\% CI [0.156, 0.579]) indicated a medium effect size.




\paragraph{Perception of the robot}
\label{sec:hypothesisH2}

To understand the difference in perception for the two systems, we compared participant ratings for warmth, competence, and discomfort provided through RoSaS. Table~\ref{tab:descriptive_stats_rosas} presents the mean and standard deviation for each scale. 
The Shapiro-Wilk test indicated that warmth ratings were not normally distributed for both systems  ($p = 0.728$ for \frameworkName, $p = 0.031$ for the baseline). We used a non-parametric Wilcoxon Signed Rank Test to analyze warmth ratings. We found no significant difference between the baseline and \frameworkName. Similarly, Competence ratings, analyzed using a mixed-effects model, showed no significant difference. Discomfort ratings were not normally distributed for either system ($p = 0.004$ for baseline, $p < 0.001$ for BT). Therefore, we used a non-parametric Wilcoxon Signed Rank Test, which showed a significant difference between the systems with lower discomfort ratings for \frameworkName (V = 483.5, $p = 0.0184$).





\paragraph{Trust and Explainability}
\label{sec:hypothesisH3}


\begin{figure}[t]
  \centering
  \includegraphics[width=0.35\textwidth]{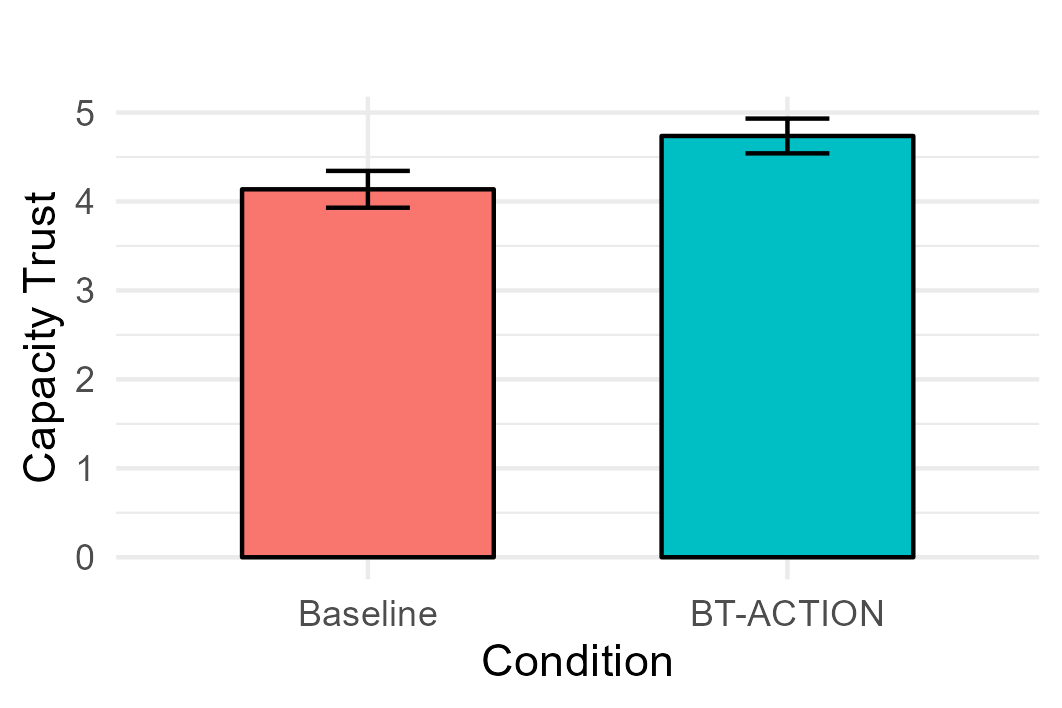}
  \vspace{-0.5em}
  \caption{Mean and standard error for MDMT Capacity Trust for the two conditions.}
  \label{fig:CapacityTrustPlot}
\end{figure}

\begin{table}[t]
\centering
\caption{Descriptive Statistics for MDMT and XAI Scales.}
\label{tab:trust_descriptive}
\begin{tabular}{lccc}
\textbf{Scale} & \textbf{System} & \textbf{Mean} & \textbf{SD} \\
\multirow{2}{*}{Capacity Trust} & Baseline & 4.3 & 0.9 \\
                                & BT & 4.8 & 0.8 \\
\multirow{2}{*}{Moral Trust}    & Baseline & 4.1 & 0.8 \\
                                & BT & 4.3 & 0.7 \\
\multirow{2}{*}{Explainability} & Baseline & 4.37 & 0.9 \\
                                & BT & 4.55 & 0.8 \\
\end{tabular}
\vspace{-2em}
\end{table}


Shapiro-Wilk tests revealed that the data for Capacity Trust were non-normal ($p = 0.045$ for the baseline system, $p = 0.129$ for \frameworkName). A Wilcoxon Signed Rank Test showed a significant difference, $V = 236, p = 0.011$, in capacity trust between the two systems illustrated in Figure~\ref{fig:CapacityTrustPlot}. \frameworkName was rated with higher Capacity Trust than the baseline (see Table \ref{tab:trust_descriptive} for descriptive statistics). A paired t-test showed no difference between the systems for Moral trust ($p = 0.055$).

The data for explainability was non-normal ($p = 0.008$ for baseline and $p = 0.005$ for \frameworkName). A Wilcoxon Signed Rank Test showed no significant differences in explainability ratings between the systems.

\paragraph{System Preference}
\label{sec:hypothesisH4}
An Exact Binomial Test revealed that the system preferences are significantly different from an assumed equal preference (\( p = 0.0022 \)). 73.2\% of participants preferred \frameworkName, while 26.8\% preferred the baseline system.

%% file: sections/05_discussion.tex
\section{DISCUSSION, CONCLUSION AND FUTURE WORK}

In this paper, we present the \frameworkName, a test-driven approach to generate a set of robot actions to fulfill user instructions. \frameworkName leverages Behavior Trees and LLMs to provide a structured, step-by-step solution for handling ambiguous, infeasible, or novel user queries. Our user study showed that \frameworkName significantly reduced task errors and enhanced user trust, leading to a clear preference over the baseline approach.

The strength of the \frameworkName framework lies in its modular and test-driven approach, which enables component-wise debugging and splitting complex queries into different categories, and it supports the adaptation of existing robot knowledge to accommodate novel situations. While the baseline prompt was consistently designed to handle similar cases, its lack of granularity increased the likelihood of errors. As a result, \frameworkName significantly reduced task errors (\textbf{H1}), and participants showed a strong preference for it (73.2\%) over the baseline approach (\textbf{H4}). These findings align with prior research emphasizing the benefits of decomposing complex tasks into manageable subtasks~\cite{meng_dcr_2024} and are consistent with users’ preference for systems that provide clear and informative feedback~\cite{lim_why_2009}. Overall, the results highlight the value of a modular understanding of user instructions, particularly in complex real-world scenarios when precise task execution and a positive user experience are fundamental, such as robotic assistance in healthcare or educational setups.


Beyond reduced error and the participants' preferences, people expressed significantly lower discomfort (\textbf{H2}) and increased trust in the robot’s capabilities (\textbf{H3}) in the \frameworkName condition. These outcomes likely stem from \frameworkName's ability to communicate the robot's capabilities and clarify ambiguities, 
aligning with prior research showing that transparent and predictable robotic behavior fosters user comfort and trust in HRI~\cite{wang_trust_2016, dogan_asking_2022, gunning_darpas_2019}. In contrast, perceptions of the robot’s competence, warmth, and moral reliability did not differ between conditions. We attribute the latter two to the nature of the interaction, which was centered on task completion rather than social engagement. This could suggest that modular structures may play a more pivotal role in task-oriented contexts than in socially driven applications, such as entertainment robots. Nonetheless, given the short duration of the interaction, future work could explore these dimensions further through longitudinal studies.

Future work can extend \frameworkName while handling infeasible comments to further detail which parts of a user's request fall outside robot capabilities. This could further increase the transparency of the decision-making process and potentially improve the user's perception of the robot.   Additionally, evaluating the framework in the wild, such as in robot cafes, could offer further valuable insights into its robustness.